\documentclass{article}
\usepackage{spconf,amsmath,graphicx}
\usepackage{algorithmic}
\usepackage{algorithm}
\usepackage{multirow}
\usepackage{amsmath,amssymb}

\title{Supplementing Missing Visions via Dialog for Scene Graph Generations}
\name{Zhenghao Zhao$^{1,*}$ \qquad Ye Zhu$^{1,*}$ \qquad  Xiaoguang Zhu$^{2}$ \qquad Yuzhang Shang$^{1}$  \qquad Yan Yan$^{1}$ \thanks{* These authors contributed equally.}}

\address{$^{1}$ Illinois Institute of Technology, USA \qquad
      $^{2}$ Shanghai Jiao Tong University, China}

\begin{document}
%
\maketitle

\begin{abstract}
Most AI systems rely on the premise that the input visual data are sufficient to achieve competitive performance in various tasks. However, the classic task setup rarely considers the challenging, yet common practical situations where the complete visual data may be inaccessible due to various reasons (\textit{e.g.}, restricted view range and occlusions). To this end, we investigate a task setting with incomplete visual input data. Specifically, we exploit the Scene Graph Generation (SGG) task with various levels of visual data missingness as input. While insufficient visual input naturally leads to performance drop, we propose to supplement the missing visions via natural language dialog interactions to better accomplish the task objective.
We design a model-agnostic Supplementary Interactive Dialog (\emph{SI-Dial}) framework that can be jointly learned with most existing models, endowing the current AI systems with the ability of question-answer interactions in natural language.
We demonstrate the feasibility of such task setting with missing visual input and the effectiveness of our proposed dialog module as the supplementary information source through extensive experiments, by achieving promising performance improvement over multiple baselines.
\end{abstract}
\begin{keywords}
Multimodal Learning, image scene understanding, language modeling
\end{keywords}
\section{Introduction}
\label{sec:intro}

The majority of the current AI systems rely on sufficient visual data (\textit{e.g.}, clear images or multiple frames from the video clip) to achieve the best performance in various vision-related tasks such as scene graph generation (SGG)~\cite{imp,motifs,graphrcnn,gu2019scene,chen2019counterfactual,li2017scene,zhang2019graphical} and visual dialog~\cite{visualdialog,de2017guesswhat,niu2019recursive,das2017learning,seo2017visual}. 
The classic experimental settings rarely consider the situations where the input visual data may be insufficient to fulfill the task objectives.
However, in practical scenarios, the missingness in the visual data is more than a common issue that can be caused by various reasons. For example, certain objects within a single image may be occluded by other objects during photography, which makes it difficult to identify their category and to infer the precise scene graph in the SGG task. Recent work~\cite{yang2021face} has also started to exploit the data privacy problems by deliberately obfuscating some sensitive information (\textit{e.g.}, human faces) from images as visual input.
To this end, we consider the computer vision task setting with insufficient visual input.

As the primary information source for various computer vision tasks, the visual input data play a significant role in most existing works to achieve competitive and promising performance. It is reasonable to expect the performance drop under the task setting with incomplete visual input. To tackle the problem, we propose to supplement the missing visual data from another information source: the natural language dialog. Intuitively, humans rely on the multi-sensory systems from various modalities (\textit{e.g.}, vision, audio, and language) to understand the surrounding world, and it is intuitive for them to ask questions about the insufficient information given a specific task to fulfill. 
To implement the proposed idea of supplementing the insufficient visual input via the natural language dialog, we introduce a model-agnostic interactive dialog framework, which can be jointly learned with most existing models and endows the models with the capability to communicate in the form of natural language question-answer interactions.
The proposed Supplementary Interactive Dialog (\emph{SI-Dial}) framework stimulate the realistic scenarios with two agents, where one agent plays the role of the actual AI system given insufficient visual input, and the other plays the role of human user or evaluator that can answer the raised questions with needed information.

\begin{figure*}[t]
    \centering
    \includegraphics[width=0.7\textwidth]{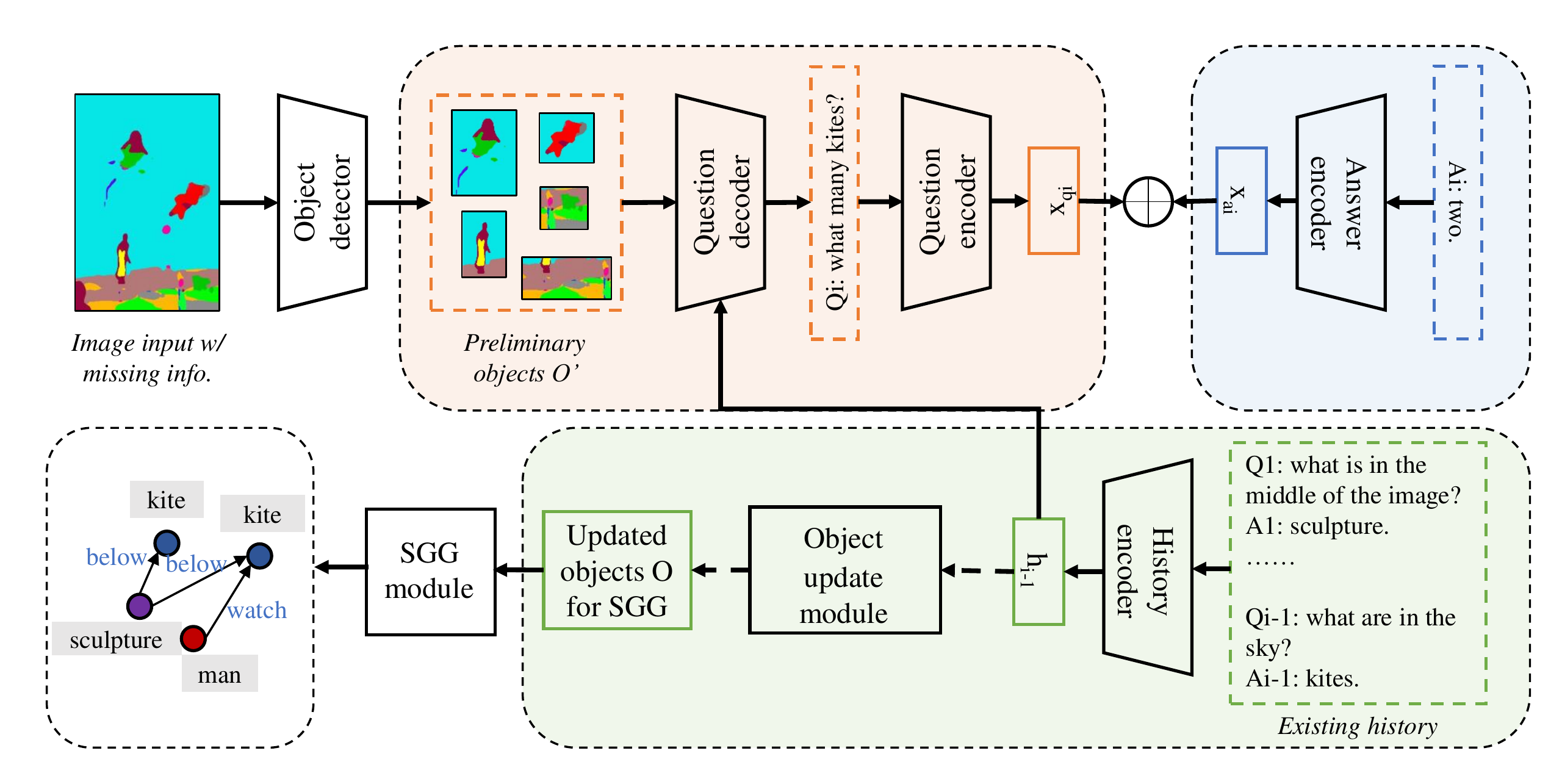}
    \vspace{-0.1in}
    \caption{The overall architecture of our proposed \emph{SI-Dial} framework. We first obtain the preliminary objects from the object detector based on the incomplete visual input, and propose to conduct an interactive dialog process. Note that the dashed lines denote the operations only after the dialog is completed) for the final scene graph generation.}
    \label{fig:model}
    \vspace{-0.2in}
\end{figure*}

 We conduct experiments on the Scene Graph Generation (SGG) task to test the feasibility of the task setting with missing visual input and to demonstrate the effectiveness of our proposed method.
 SGG task aims to generate a graphical representation of the scene from given images.
 Specifically, we pre-process the visual data to provide three different levels of missingness: obfuscations on the objects (\textit{e.g.}, humans, cars), obfuscations on the entire images, and the semantically masked visual images.
 The masked visual data has more severe missingness compared to the other two levels.
 We then design the dialog formulation by allowing the model to ask natural language questions and then provide the answers to the raised questions. 
 Specifically, different from most existing works in the field of visual dialog~\cite{visualdialog,avsd-alamri2019audio}, which concentrate on answering visual context-related natural language questions~\cite{antol2015vqa,vqa7,vqa5,vqa3}, our dialog design emphasizes the abilities of AI systems to ask informative questions about missing visions. 
 The experimental results show promising performance improvement with our proposed framework compared to multiple baselines.
 Notably, similar to the findings from~\cite{yang2021face} where the face obfuscated images only cause trivial performance drop for classifications and object detection, we also observe empirical evidence that not all levels of vision missingness will severely hinder the experimental results, which implies the potential redundancy of visual information for classic task setup and brings the insights for future research directions to establish and develop more reliable and privacy-preserved AI systems. 
 
The contributions of this paper are:
1) We investigate a novel SGG task setting with insufficient visual input data, and propose to supplement the missing information from the natural language dialog.
2) We propose a model-agnostic dialog framework, \emph{SI-Dial}, which can be jointly trained with various existing models and endows the AI systems with the interactive communication abilities.
3) We perform extensive experiments and analysis with insufficient visual input in three different levels of data missingness, and demonstrate that our proposed framework can effectively provide supplementary information from the dialog interactions and achieve considerably better performance compared to multiple baselines. In the meanwhile, we empirically prove that not all levels of missing visions leads to severe performance drop, which brings insights for further research directions.

\vspace{-0.1in}
\section{Methodology}

\vspace{-0.1in}
\subsection{Task Formulations}

For the SGG task, let $I_{m}$ to be the image input with various levels of data missingness. The overall goal is to establish a model to generate a scene graph $\mathcal{G}$ from the given image input $I_{m}$.
A scene graph mainly consists of a set of objects $O = \{o_1, o_2, ..., o_n\}$ with each object $o_i$ assigned to a certain class label $C$, a corresponding set of bounding boxes $B = \{b_1, b_2, ..., b_n\}$ with $b_i \in \mathbb{R}^4$, and a set of relationship $R = \{r_1, r_2, ..., r_m\}$ with each $r_i$ representing a predicate between a pair of objects.
Typically, the nodes of a scene graph are represented by the objects, while the edges are represented by the relations between object pairs.
Our proposed method factorizes the entire pipeline into three parts:
\begin{equation}
\centering
    P(G|I_{m}) = P(B,O'|I_{m})P(O|O',QA)P(R|B,O),
\end{equation}
\noindent where $O'$ denotes the preliminary objects roughly detected from the incomplete visual input and will be updated during the dialog interactions. $QA=\{qa_1, qa_2, ..., qa_{N_R}\}$ represents $N_R$ question-answer pairs given the initial visual input.

For the first step of object detection $P(B,O'|I_{m})$, we follow the previous implementations as in~\cite{imp,graphrcnn,motifs,tang2018learning}, where a standard object detection model is adopted to extract the visual representations for $n$ objects and its corresponding bounding box regions.
For the second step of dialog interactions $P(O|O',QA)$, we deploy our proposed \emph{SI-Dial} framework to update the preliminary representations extracted from the detector, which we describe in details in Section~\ref{subsec:si-dial}.
For the third step of scene graph generation $P(R|B,O)$, we test multiple popular existing SGG methods, including the state-of-the-art models, to show the effectiveness of the proposed framework compared to the case $P(R|B,O')$ in Section~\ref{sec:exp}.

\vspace{-0.1in}
\subsection{SI-Dial for Missing Visions}
\label{subsec:si-dial}


We formulate the dialog with $N_R$ rounds of QA interactions. Specifically, given the visual input data with partially missing visions, the AI system is given $N_R$ chances to ask questions based on the visual input and existing dialog history. We then provide answers to the raised questions. After the dialog is completed, the dialog with $N_R$ QA pairs are used as the supplementary information for the initial visual input. 

Overall speaking, our proposed \emph{SI-Dial} takes the preliminary object representations (\textit{i.e.}, node and edge features from the object detector) as input, and outputs the updated representations with supplementary information incorporated from the dialog interactions: 
\begin{equation}
\centering
  Input:{O'=\{V',E'\}} \Rightarrow  Output:{O=\{V,E\}},
\end{equation}
\noindent where the nodes $V'$ are typically represented by the visual features extracted from the region proposals, and the edges $E'$ are denoted by the visual features extracted from the unions of region proposals between pairs of objects. 
Since the dialog process includes multiple rounds of question-answer interactions, our explanations below are based on $i$-th round of QA. During the experiments, we set the number of QA rounds $N_R$ to be 10.

\noindent \textbf{Question Encoder.}
Similar to previous visual dialog works~\cite{visualdialog,zhu2021saying}, we adopt the discriminative dialog generation setting where the questions raised are generated by selecting the most appropriate one from the given candidates. We adopt the Sentence-BERT~\cite{reimers-2019-sentence-bert,reimers-2020-multilingual-sentence-bert} to encode and extract the question embedding for all the given question candidates. 
\begin{equation}
  x_j = QE(q_{cand., j}),\: j \in \{1,2,...,N_{cand.}\}, 
\end{equation}
\noindent where $x_j$ denotes the $j$-th question embedding, $QE$ represents the question encoder, $q_{cand.,j}$ is the $j$-th question, $N_{cand.}$ is the total number of question candidates.

\noindent \textbf{Question Decoder.}
The question decoder aims to generate and select the $i$-th question based on the preliminary incomplete visual input $O'$ and the existing dialog history with $i$-1 rounds of QA pairs. Specifically, it consists of a two-layer convolutional network with a residual block that fuses the dialog history embedding and preliminary visual objects, and then converts the fused features to the question embedding for computing similarity scores. Next, the question decoder selects the question that has the highest similarity score with the generated question embedding.
\begin{equation}
  q_i = argmin_{k}\: Sim.(QD(O', x_{his,i-1}), x_j), 
\end{equation}
\noindent where $q_i$ is the raised question for this $i$-th round, $QD$ represents the question decoder, $x_{his,i-1}$ is the existing dialog history with $i-1$ QA pairs. The answer $a_i$ corresponding to the raised question $q_i$ is provided and encoded following the similar way as for the question decoder. The $i$-th QA pair is therefore obtained by fusing the question and answer embeddings $x_{qa_i} = (x_{q_i},x_{a_i})$.

\noindent \textbf{History Encoder.}
The history encoder is for interactively encoding the QA pairs from the dialog. Specifically, in order to emphasize the information from the newly generated QA pair $qa_{i}$ from the $i$-th round, we adopt a similar technique as in~\cite{zhu2020describing}, which dynamically encodes the current QA pair into the history encoder. Concretely, the history encoder takes the existing dialog history $x_{his,i-1}$ and the newly generated QA pair $x_{qa_i}$ as input, and fuses the two input in a way that the new pair always maintains the same dimension as the existing history. The history encoder consists of a two-layer convolutional network with a residual block.
The final output $x_{his,N_R}$ from the history encoder is used as the supplementary information for the missing visual input.
\begin{equation}
    x_{his,i} = HE(x_{his,i-1},x_{qa_i}), 
\end{equation}
\noindent where $x_{his,i}$ is the dialog with $i$ QA pairs. When $i= N_R$, $x_{his,N_R}$ is the final output of our \emph{SI-Dial} for supplementing the initial visual input.

\noindent \textbf{Vision Update Module.}
Having obtained the interactive dialog $x_{his,N_R}$ as the supplementary source to missing visual input, we update the preliminary objects $O'$ obtained from the incomplete visions by incorporating the dialog information. The vision update module takes the preliminary object features and the entire dialog history features as input, and outputs the updated object features $O$ for scene graph generation. Specifically, the vision update module adopts the cross-modal attention network adapted from~\cite{schwartz2019factor}.

\begin{table*}[t]
\centering
\scalebox{0.79}{
\begin{tabular}{c|c|ccc|ccc|ccc}
\hline
 &        &  \multicolumn{3}{c|}{Predicate Classification} & \multicolumn{3}{c|}{Scene Graph Classification} & \multicolumn{3}{c}{Scene Graph Detection} \\ \hline
Vision Input & Model   & mR@20         & mR@50         & mR@100        & mR@20          & mR@50         & mR@100         & mR@20        & mR@50        & mR@100       \\ \hline \hline
\multirow{5}{*}{Original VG} & IMP$^{\dagger}$   &  -       &    9.8   &  10.5   &  - &   5.8   &  6.0    &    -     &     3.8     &    4.8      \\  
 & FREQ.$^{\dagger}$  & 8.3    &      13.0       &  16.0 &    5.1          & 7.2             &   \textbf{8.5}   &  \textbf{4.5}     &     \textbf{6.1}         &         \textbf{7.1}       \\ 
 & MOTIF$^{\dagger}$      &     11.5          &    14.6 &          15.8     &     \textbf{6.5}          &   \textbf{8.0}  &    \textbf{8.5}     &         4.1       &     5.5         & 6.8         \\  
 & VCTREE$^{\dagger}$ &      \textbf{11.7}         &    \textbf{14.9}  &    \textbf{16.1}         &      6.2         &          7.5     &       7.9         &       4.2       &     5.7         &      6.9        \\ 
 \hline \hline
\multirow{6}{*}{Object Blur}   

  & MOTIF     &      11.23        &   14.36            &       15.71        &     6.20       &      7.51         &      7.90        &    4.13       &   5.48      &       6.82       \\  
 & VCTREE   &    11.48          &      14.61         &      15.88         &      5.97         &        7.46      &   7.85            &      4.24        &   5.66          &   6.93           \\ 
 & MOTIF+Random QA &  11.52 & 14.78 & 16.08 &  5.82  & 7.34  &7.86 & 4.64 & 6.18 & 7.24 \\
 & VCTREE+Random QA & 11.55 & 14.83 & 16.17 &  5.51 & 7.03  & 7.48 & 4.71 & 6.23 & 7.43\\
  & MOTIF+SI-Dial & 13.31 & 16.74 & 18.09 & \textbf{6.96}  & \textbf{8.51}  & \textbf{9.00} & 5.77 & 7.76 & 9.12\\
  & VCTREE+SI-Dial & \textbf{13.40} & \textbf{16.88} & \textbf{18.26} & 6.67  & 8.12  & 8.59 & \textbf{5.88} & \textbf{7.92} & \textbf{9.28} \\
 \hline \hline
 
 \multirow{6}{*}{Image Blur}   

  & MOTIF     &      11.69       &   14.31            &     15.64        &      6.24        &      7.56         &       7.90         &     3.88         &      5.21       &     6.37       \\  
 & VCTREE   &    11.72          &     14.38        &     15.78        &     6.03          &     7.39          &     7.83         &       3.82       &   5.18          &     6.33        \\ 
 & MOTIF+Random QA & 11.57  & 13.93 & 15.14 & 6.75  & 8.21  & 8.69  & 4.10 & 5.39 & 6.26 \\
 & VCTREE+Random QA & 11.70 & 14.26 & 15.53 &  6.68 & 8.03&  8.55  & 4.07 & 5.34 & 6.25 \\
  & MOTIF+SI-Dial & 12.90 & 16.26 & 17.91 & \textbf{8.41}  & \textbf{10.33}  & \textbf{11.00} & 5.05 & 6.96 & \textbf{8.23}\\
  & VCTREE+SI-Dial & \textbf{13.62} & \textbf{17.18} & \textbf{18.49} & 7.93  & 10.02  & 10.86 & \textbf{5.24} & \textbf{7.08}  & 8.11 \\
 \hline \hline
 \multirow{6}{*}{Semantic Masked}   

  & MOTIF     &       11.61        &   14.28           &       15.57        &       4.45         &      5.41         &       5.68         &     2.80         &       3.89       &       4.76       \\  
 & VCTREE   &     11.68          &      14.32       &     15.59         &      4.40  &     5.38          &      5.69          &     2.80         &        3.87      &      4.68        \\ 
 & MOTIF+Random QA & 12.00 & 15.32 & 16.67 & 5.83  & 7.14  & 7.65 & 2.79 & 3.86 & 4.68 \\
 & VCTREE+Random QA & 12.28 & 15.69 & 17.04 &  5.66 & 7.01  & 7.28 & 2.92 &4.01 &4.85 \\
  & MOTIF+SI-Dial & \textbf{12.79} & 16.26 & 17.58  & \textbf{6.44} & \textbf{7.85}  & \textbf{8.33} &3.03 & 4.21& 4.92\\
  & VCTREE+SI-Dial & 12.73 & \textbf{16.35} & \textbf{17.63} &  6.21 & 7.68  & 8.05& \textbf{3.15} & \textbf{4.28} &  \textbf{5.00}\\
 \hline \hline
\end{tabular}}
\caption{Quantitative evaluations for the SGG with missing visions. The results are reported on mean Recall. 
}
\label{tab:sggresult}
\vspace{-0.2in}
\end{table*}

\vspace{-0.1in}
\subsection{Pipeline Training}
\vspace{-0.05in}
We train the entire pipeline following the widely adopted stepwise training mechanism as in previous studies~\cite{imp,motifs,tang2018learning,tang2020unbiased,graphrcnn}. We firstly train the object detector on the image input with missingness. For the second stage, we freeze the parameters in the objector detector and attach the proposed \emph{SI-Dial} to the pipeline of the existing SGG models and train it jointly with the SGG module using cross-entropy loss for the final objects and predicates predictions. 

\vspace{-0.1in}
\section{Experiments}
\label{sec:exp}
\vspace{-0.1in}

\subsection{Experimental Setup}

\noindent \textbf{Datasets.}
We use the benchmark Visual Genome (VG) dataset~\cite{vgdataset} for experiments. The VG dataset contains in total 108,077 images and 1,445,322 question-answer pairs. 
We firstly perform the vision data processing to obtain three levels of missingness: the obfuscations applied on the objects, the obfuscations applied on entire images, and the semantically masked images.

\noindent \textbf{Question Candidates.}
We provide 100 question candidates for the model to select from for each given image, similar to the work of visual dialog~\cite{visualdialog}.
Note that the ground truth question-answer pairs from the dataset annotations are not evenly distributed, meaning some of the images do not have corresponding questions and answers, therefore, the 100 candidates are formed from two sources. For the images with GT QA pairs, we include the GT pairs as part of the candidates, and the rest are randomly sampled from the dataset; while for the rest images without GT QA pairs, all the candidates are randomly sampled from the dataset.

\noindent \textbf{Evaluations.}
We evaluate our generated scene graphs using the three evaluation metrics: (1) \textit{Predicate Classification (PredCls)}: predict the predicates (relations) given the sets of ground truth bounding boxes and object labels. (2) \textit{Scene Graph Classification (SGCls)}: predict the predicate as well as the object labels given the sets of ground truth bounding boxes. (3) \textit{Scene Graph Detection (SGDet)}: predict the bounding boxes, the predicates as well as the object labels. 
We calculate and report the mean recall@K scores for the above metrics in experiments.

\vspace{-0.1in}
\subsection{Scene Graph Generation Results}

\noindent \textbf{Original VG.}
We present the SGG results on the original VG dataset without missing information in the first group of Table~\ref{tab:sggresult} for comparisons. These quantitative results are reported by several most popular existing SGG models include the IMP~\cite{imp}, FREQ~\cite{motifs}, MOTIF~\cite{motifs,tang2020unbiased}, and VCTREE~\cite{tang2018learning,tang2020unbiased}. 
IMP and MOTIF introduce the feature interaction of local and global feature context information.
FREQ predicts object labels using a pretrained detector. VCTREE proposes structure designed to learn various types of contextual information.

\noindent \textbf{Baselines.}
The baselines are established using the incomplete visual input without the supplementary information from the natural language dialog. 
We mainly report the results obtained using the MOTIF~\cite{motifs} and VCTREE~\cite{tang2018learning,tang2020unbiased} models. 

We observe that the \textit{PredCls} does not fluctuate much in case of missing visions compared to other two metrics \textit{SGCls} and \textbf{SGDet}. It is consistent with the previous findings in~\cite{motifs}, where the authors find that the object labels are highly predictive of relation labels but not vice-versa. In contrast, \textit{SGCls} and \textit{SGDet} drops more or less with missing visions. Also, it is worth noting that for the datasets with object obfuscations and image obfuscations do not cause a severe performance drop, which implies that the setting with original images may contain the visual redundancy for the SGG task. However, the semantic masked images evidently impair the performance as we expected, which is also reasonable due to the reason that this preprocessing causes most vision missingness among three levels.

\noindent \textbf{SI-Dial.}
The results for the generated scene graphs from incomplete images incorporated with the proposed \emph{SI-Dial} are also presented in Table~\ref{tab:sggresult}. We observe that dialog can serve as an effective supplementary information source in case of missing visions, especially for the semantically masked visual input with most severe missingness level. 
We also compare the proposed method against the random QA case. The SGG results do not improve in an obvious way, or even get hindered as for the \textit{PredCls} and \textbf{SGCls} on the object obfuscation and image obfuscation datasets. The comparisons indicate that our propose \emph{SI-Dial} framework indeed learns to ask questions in a meaningful way. 
In addition, we also show again as in the baseline situations, that the first two levels of visual missingness, are innocuous for the SGG tasks. It provides empirical evidence and further insights to bring deliberately obfuscations for privacy concerns as in~\cite{yang2021face}.  

\vspace{-0.2in}
\section{Conclusion}
\vspace{-0.1in}
In this paper, we investigate the SGG task setting with missing input visions, and propose to supplement the missing visual information via interactive natural language dialog using the proposed \emph{SI-Dial} framework. Extensive experiments on the benchmark dataset with various levels of missingness demonstrate the feasibility of the task setting and the effectiveness of our proposed model by achieving promising performance improvement. 
We plan to expand our approach on tasks that require the integration of missing visual data with natural language dialogue interactions in our future works.

\noindent \textbf{Acknowledgements:} This work is supported by NSF IIS-2309073 and NSF SCH-2123521. This article solely reflects the opinions and conclusions of its authors and not the funding agency.


\bibliographystyle{IEEEbib}
\bibliography{refs}

\end{document}